\PassOptionsToPackage{breaklinks, colorlinks}{hyperref}
\documentclass[11pt]{article}

\usepackage[final]{acl}
\usepackage{fontspec}
\usepackage{polyglossia}
\usepackage{times}

\usepackage[T1]{fontenc}
\usepackage{array}
\usepackage{enumitem}
\setlist[itemize]{noitemsep, nolistsep}
\newfontfamily\devanagarifont{NotoSansDevanagari.ttf}[Script=Devanagari]
\newcommand{\devanagari}[1]{{\devanagarifont #1}}

\usepackage{microtype}

\usepackage{times}
\usepackage{microtype}

\usepackage{inconsolata}

\usepackage{graphicx}
\usepackage{booktabs} 
\usepackage{pifont}
\definecolor{mediumgreen}{HTML}{13B015}
\newcommand{\tick}{{\color{mediumgreen} \ding{51}}}
\newcommand{\cross}{{\color{red} -}}
\usepackage{amsmath}
\usepackage{amssymb}
\usepackage[linesnumbered,ruled,vlined]{algorithm2e}
\usepackage{adjustbox}
\usepackage[breaklinks, colorlinks]{hyperref}

\usepackage{xcolor}

\newcommand{\chandomitra}{{\texttt{Chandomitra}}}
\newcommand{\emb}[1]{\texttt{emb}(#1)}

\usepackage{tikz,lipsum}
\usepackage[most]{tcolorbox}
\usepackage{etoolbox}
\AtBeginEnvironment{tcolorbox}{\small}
\usepackage{authblk}

\title{Pingala: Prosody-Aware Decoding for Sanskrit Poetry Generation}

\author{Manoj Balaji Jagadeeshan$^{1}$$^{*}$ Atul Singh$^{1}$$^{*}$  Nallani Chakravartula Sahith$^{1}$  Amrith Krishna$^{2}$  Pawan Goyal}
\affil[1]{Indian Institute of Technology, Kharagpur}
\affil[2]{Oriflow}

\begin{document}
\maketitle
\def\thefootnote{*}\footnotetext{These authors contributed equally to this work}\def\thefootnote{\arabic{footnote}}

\begin{abstract}
Poetry generation in Sanskrit typically requires the verse to be semantically coherent and adhere to strict prosodic rules. In Sanskrit prosody, every line of a verse is typically a fixed length sequence of syllables adhering to prescribed binary patterns of syllable weights.  We observe that instead of treating a verse as a monolithic sequence, segmenting them as grouped-lines leads to significant improvement in semantic coherence by 10\% with comparable metrical adherence. Specifically, Pingala, our proposed decoding approach is designed to encourage every line to have well-formed words and our token selection biases the model towards it by preferring longer tokens. Writing in Sanskrit follows phonemic orthography, hence using a phonetically aware transliteration scheme, SLP1,  increased the metrical alignment by 46\% with comparable semantic similarity, for a instruction fine-tuned large language models like Phi-4. We also introduce a new approach for reference-free evaluation using cross-encoders which achieved better alignment with true poetry instances.
\end{abstract}
\section{Introduction}

Prosodic constraints in Sanskrit are articulated through fixed-length, syllable-level binary patterns, composed of light (laghu) and heavy (guru) units. The generation of Sanskrit poetry frequently necessitates the simultaneous fulfilment of both semantic and prosodic constraints, a requirement that often involves a degree of compromise. A Sanskrit poet may prioritize the arrangement of words to satisfy these metrical constraints, potentially at the expense of verbal cognition. Consequently, any computational methodology for Sanskrit poetry generation must ensure both linguistic veracity and stringent adherence to metrical rules, rendering this endeavour especially demanding for neural text generation systems.

\chandomitra$\:$\citep{jagadeeshan-etal-2026-chandomitra} recently introduced a comprehensive dataset and benchmark for generating Sanskrit Anu\d{s}\d{t}ubh poetry. This seminal work offers a large English $–$ Sanskrit parallel corpus and evaluates diverse language models, including encoder–decoder and decoder-only architectures. The publication also proposes a reference-free semantic evaluation metric using multilingual sentence embeddings to assess coherence. Critically, the benchmark explores constrained decoding strategies to ensure strict metrical adherence. Results show that constrained decoding significantly improves metrical correctness across all models, highlighting a trade-off between prosodic compliance and semantic fidelity in Sanskrit verse generation.


However, the proposed reference-free metric and the constrained-decoding approach reveal important limitations. The semantic metric - despite being a useful starting point - yields a maximum score of 73.45\% against ground truth while negative examples average 50.97\%, a narrow spread of $\approx$22.48 percentage points, which indicates limited discriminative power between high - and low-quality outputs. Similarly, constrained decoding achieves strong metrical compliance but scores poorly on human judgments of semantic coherence (Avg. 2.13), implying that enforcing strict syllabic patterns during generation often degrades meaning. These findings expose a persistent trade-off between prosodic fidelity and semantic adequacy and highlight clear room for improved evaluation and decoding strategies that balance meter and meaning.

To address these limitations, we propose a the following: \\\textbf{(1)} we introduce, \texttt{Pingala}, a decoding-time guidance mechanism, defined as a shaping function based on metrical milestones - biases token selection toward candidates that are longer and that which complete well-formed words at p\={a}da boundaries, operationalizing the yati principle of Anuṣṭubh prosody. This ensures that the generated candidates adhere to the structural rules of the \textit{Anu\d{s}\d{t}ubh} meter without requiring expensive retraining.\\\textbf{(2)} To overcome the representational bottlenecks of Bi-Encoders, we employ a Cross-Encoder architecture that processes the source and candidate jointly, enabling full cross-attention. This allows for a far more granular assessment of lexical and syntactic fidelity, effectively filtering out high-confidence errors that escape standard retrieval models \cite{mesgar-etal-2023-devil}.\\\textbf{(3)} \noindent While all training and evaluation data in this work are natively represented in Devanagari\footnote{\url{https://en.wikipedia.org/wiki/Devanagari}}, we observe a large and consistent gain when instruction fine-tuning Phi-4 on Anu\d{s}\d{t}ubh samples encoded in the SLP1\footnote{\url{https://en.wikipedia.org/wiki/SLP1}} transliteration scheme and decoding with \texttt{Pingala} (Full \% = 95.92, Partial \% = 98.31). Prior work in multilingual and low-resource NLP has shown that representation and tokenization choices strongly influence downstream performance: romanization reduces vocabulary fragmentation and script mismatch with predominantly Latin-trained tokenizers \citep{limisiewicz2023tokenization,purkayastha2023romanization}, while phonemic ASCII encodings improve normalization stability and facilitate the learning of sound-level regularities \citep{adiga2021asr}. Recent Sanskrit-specific modeling studies further demonstrate that script and encoding decisions materially affect performance on morphology- and structure-sensitive tasks \citep{nehrdich2024byt5sanskrit}. Together, these results motivate our third contribution: using SLP1 transliteration as a preprocessing and representation choice significantly amplifies the effectiveness of instruction fine-tuning when combined with metric-aware decoding, yielding near-perfect metrical conformity for Sanskrit poetry generation.\\
We evaluate our approach using both automated metrics and expert human evaluation, comparing it against strong baselines including NLLB and Phi-4. Our results demonstrate that the proposed method achieves near-perfect metrical compliance while significantly improving semantic alignment. Specifically, we show that introducing our prosody defined metrical weight, restores full syntactic correctness (100\% metric satisfaction in ablation studies) without degrading semantic quality, provided the weight is appropriately tuned. Furthermore, our human evaluation highlights the superior discriminative power of the Cross-Encoder over the Bi-Encoder, validating its necessity for high-precision tasks in low-resource computational philology.

\paragraph{Problem Statement} Formally, given an English input sentence (prose) x, the objective is to generate its corresponding Sanskrit poetic verse, \^{y}, which adheres to the syntactic constraints of the Anu\d{s}\d{t}ubh chanda while faithfully preserving the meaning of x. The corresponding ground-truth Sanskrit poem, when available, is denoted by y.

In essence, this task simultaneously involves machine translation and structured poetry generation, requiring the output to be both semantically accurate and metrically compliant.

\section{Reference Free Evaluation}
Every problem statement requires a robust evaluation mechanism. For the task of structured poetry generation from input prose language, and the task being cross-lingual in nature, we have to evaluate the poetry on two aspects (1) Does the generated poetry actually follow the expected style(in this case the chandas) (2) is the generated poetry, semantically faithful with respect to the given input sentence?. Thus, to answer them, we define the Syntactic Metrics(\autoref{sec:syntactic-metric}) and Semantic Metrics(\autoref{sec:semantic_similarity}).
\subsection{Syntactic Metrics}
\label{sec:syntactic-metric}
Following \chandomitra, we evaluate metrical correctness using two metrics. \textbf{Full Anu\d{s}\d{t}ubh} measures the proportion of generated verses that strictly satisfy all Anu\d{s}\d{t}ubh metrical constraints, computed using \texttt{skrutable} \cite{skrutable}. \textbf{Partial Anu\d{s}\d{t}ubh} measures the proportion of outputs with the required 32-syllable length, providing a relaxed upper bound that subsumes the full-metric scores. 
\subsection{Semantic Similarity}
\label{sec:semantic_similarity}
We employ, infoxlm\footnote{\url{https://huggingface.co/microsoft/infoxlm-large}} \cite{chi-etal-2021-infoxlm}, a cross-encoder architecture for comparing translations, as it jointly processes both the source and translated texts through full self-attention, thereby capturing fine-grained semantic interactions between them. This approach contrasts with bi-encoder methods, also used in \chandomitra - the BGE-M3\footnote{\url{https://huggingface.co/BAAI/bge-m3}} \cite{bge-m3} model fine-tuned on \texttt{Mitrasa\d{m}graha} \cite{nehrdich2026mitrasamgrahacomprehensiveclassicalsanskrit}, which independently map texts into a shared embedding space and measure similarity via cosine distance — an approach known to suffer from representation degeneration and anisotropic embedding distributions \cite{gao2018representation, rajaee2022isotropy}. Prior work has consistently demonstrated that cross-encoders achieve higher accuracy than bi-encoders for pairwise text comparison tasks \cite{reimers-gurevych-2019-sentence, Humeau2020Poly-encoders:, rosa2022defense}, including semantic answer similarity where cross-encoders correlate most strongly with human judgment \cite{risch-etal-2021-semantic}. Similarly, state-of-the-art translation evaluation metrics such as COMET \cite{rei-etal-2020-comet} and BLEURT \cite{sellam-etal-2020-bleurt} leverage joint encoding of source and translation, further validating that direct cross-attention between text pairs captures translation quality more faithfully than independent embedding comparison.\\
\begin{table}[t]
    \centering
    \resizebox{0.482\textwidth}{!}{
    \begin{tabular}{lccc}
    \toprule
        \textbf{Eval Metric} &  \textbf{Ground Truth}&  \textbf{Pingala}& \textbf{CD}\\
        \midrule
         Semantic Similarity$^{*}$&  73.46& 69.05 &64.96 \\
         Human Ratings&  \cross& 2.70 & 2.13\\
         \bottomrule
    \end{tabular}
    }
    \caption{Comparison of semantic similarity scores and human evaluation ratings( on standard constrained decoding(CD) and Pingala outputs. The bi-encoder assigns similar scores to translations with vastly different human-perceived quality. $^{*}$as used in \chandomitra. ~Human ratings for Pingala is average of the ratings specified in \autoref{tab:semantic-human} while for the CD is from \chandomitra.}
    \label{tab:semantic-human}
\end{table}
\noindent\textbf{Empirical motivation: }\autoref{tab:semantic-human} illustrates the problem: the bi-encoder returns an average similarity of roughly \(73.5\%\) on ground-truth translation pairs while human raters assign near-perfect judgments. Conversely, candidate outputs that humans rate poorly still receive deceptively high bi-encoder scores. This numerical/human mismatch indicates poor separation and motivates the cross-encoder replacement. 

\subsubsection{Implementation notes.} For training the cross-encoder we use a binary classification objective as defined in \autoref{sec:crossenc-loss} over (English, Sanskrit) pairs, with a mixture of gold positives and mined hard negatives (including metre-preserving but semantically-different candidates) as defined in \autoref{sec:crossenc-data}. Model checkpoints are selected to explicitly maximize $G(\theta)$ as defined in \autoref{sec:crossenc-opt}, on the validation set. We found that this curriculum improves the performance of the model.

\subsubsection{Cross-Encoder Model Definition}

Let $x = (e,s)$ denote the concatenated input sequence $[\text{English};\ \text{Sanskrit}]$. The cross-encoder is a parameterized function
\[
g_{\theta} : \mathcal{X} \rightarrow \mathbb{R},
\]
which maps the joint input to a scalar logit
\[
z = g_{\theta}(x).
\]
The predicted probability that $s$ is a correct translation of $e$ is obtained via the sigmoid function:
\[
p_{\theta}(x) = \sigma(z) = \frac{1}{1 + e^{-z}}.
\]

\subsubsection{Training Data Construction}
\label{sec:crossenc-data}

We begin with the dataset of parallel translations as used by \texttt{mitrasa\d{m}graha}:
\[
\mathcal{D}_{+} = \{(e_i, s_i)\}_{i=1}^{N}.
\]
Negative examples are generated in two stages:
\begin{itemize}[noitemsep]
  \item \textbf{Easy negatives} $\mathcal{D}_{-}$, constructed by pairing an English sentence with a randomly selected non-corresponding Sanskrit sentence.
  \item \textbf{Hard negatives} $\mathcal{D}_{\mathrm{hard}}$, obtained by retrieving semantically similar but incorrect Sanskrit candidates using an embedding-based retrieval model.
\end{itemize}

The final training set is
\[
\mathcal{D} = \mathcal{D}_{+} \cup \mathcal{D}_{-} \cup \mathcal{D}_{\mathrm{hard}},
\]
where each example is labeled with $y \in \{0,1\}$.

\subsubsection{Weighted Binary Cross-Entropy Objective}
\label{sec:crossenc-loss}

We train the cross-encoder using a class-weighted binary cross-entropy loss:
\begin{equation}
\begin{split}
    \mathcal{L}(\theta) =
-\frac{1}{|\mathcal{D}|}
\sum_{(x,y)\in\mathcal{D}}
w(y)\Big[
y\log p_{\theta}(x) \\ + (1-y)\log\big(1-p_{\theta}(x)\big)
\Big],
\end{split}
\end{equation}
where $w(1)=\alpha>1$ and $w(0)=1$. The positive-class weight $\alpha$ biases training toward preserving high confidence on correct translations while suppressing false positives.

The Rationale for this is available at \autoref{appendix:class-weighted}.

\subsubsection{Probability Gap as the Optimization Target}
\label{sec:crossenc-opt}
We define the \emph{probability gap} as:
\[
G(\theta) =
\mathbb{E}_{x\sim\mathcal{D}_{+}}[p_{\theta}(x)]
-
\mathbb{E}_{x\sim\mathcal{D}_{\mathrm{cross}}}[p_{\theta}(x)],
\]
where $\mathcal{D}_{\mathrm{cross}}$ denotes cross-translation (incorrect) pairs used for evaluation.

\subsubsection{Empirical Outcome}

Using the proposed cross-encoder with curriculum hard negatives and weighted optimization, the probability gap increased from approximately $22\%$ under the bi-encoder formulation to approximately $88\%$, refer to \autoref{tab:semantic-score-spread}, demonstrating a decisive improvement in translation verification accuracy as seen in \autoref{tab:semantic-acc}. The thoeritical rationale on why spread is high is available at \autoref{appendix:spread-ratonale}

\begin{table}
    \centering
    \resizebox{0.482\textwidth}{!}{
    \begin{tabular}{cccc}
    \toprule
         \textbf{Approach}&  \textbf{Positive}&  \textbf{Negative(Avg.)}& \textbf{Spread}\\
    \midrule
       bge-m3  & 73.46\% & 50.97\% & 22\%\\
        cross & 90.02\% & 1.67\% & 88.33\%\\
    \bottomrule
    \end{tabular}
    }
    \caption{Semantic similarity scores based on the cross-encoder}
    \label{tab:semantic-score-spread}
\end{table}

\begin{table}
    \centering
    \begin{tabular}{ccc}
    \toprule
    \multicolumn{3}{c}{\textbf{Mean Acc.}} \\
         \textbf{Batch Size}&  \textbf{bge-m3}& \textbf{cross-encoder}\\
         \midrule
         Full&  46.78\%& 74.74\%\\
         16&  82.35\%& 84.71\%\\
         4&  91.53\%& 92.87\%\\
         \bottomrule
    \end{tabular}
    \caption{Mean Accuracy: bi-encoder model(\chandomitra) vs proposed cross-encoder model}
    \label{tab:semantic-acc}
\end{table}

\subsubsection{Qualitative Analysis}
\label{sec:human_eval}

\begin{table}[ht]
    \centering
    \resizebox{0.48\textwidth}{!}{
    \begin{tabular}{lcccc}
    \toprule
    \textbf{Model} & \multicolumn{2}{c}{\textbf{Raw}} & \multicolumn{2}{c}{\textbf{Isotonic Calibrated}} \\
    & Pearson & Spearman & Pearson & Spearman \\
    \midrule
    Bi-Encoder   & 0.6075 & 0.6094 & 0.6559 & 0.6459 \\
    Cross-Encoder& 0.6767 & 0.5833 & 0.7248 & 0.7175 \\
    \bottomrule
    \end{tabular}
    }
    \caption{Correlation between human semantic ratings (N=200) and automatic scorers.}
    \label{tab:human_corr}
\end{table}

From \autoref{tab:human_corr}, the cross-encoder shows higher agreement with the expert rater than the bi-encoder, and isotonic calibration further improves both Pearson and Spearman correlations. In practice, the combination \textbf{cross-encoder + isotonic calibration} yields the best alignment with human judgments and is therefore used for final candidate ranking in our pipeline.
Details on how qualitative analysis was performed is available at \autoref{appendix:qualitative-ce}

\section{Method}
\label{sec:method}
Our method combines metrical fine-tuning with a prosody-aware decoding procedure. The two core ideas are: (1) shift the model's token priors so metrically-useful continuations become more probable - achieved through fine-tuning and (2) decoding towards structurally stable prefixes using a sparse re-weighting function applied to logits during inference. We also investigate SLP1 transliteration as an alternative to Devanagari in \autoref{sec:method:slp1}.

\subsection{Fine-tuning}
We fine-tune NLLB-dist-1.3B \cite{nllbteam2022languageleftbehindscaling}, and instruction fine-tune Phi-4 \cite{abdin2024phi4technicalreport} using the prompt(\autoref{image/IFT-prompt}) in \autoref{sec:prompt}, on the train dataset split of \chandomitra; which consists of 8306 English Prose - Sanskrit Poetry parallel data and test split consists of 1421.
\begin{figure*}
    \centering
    \includegraphics[width=\linewidth]{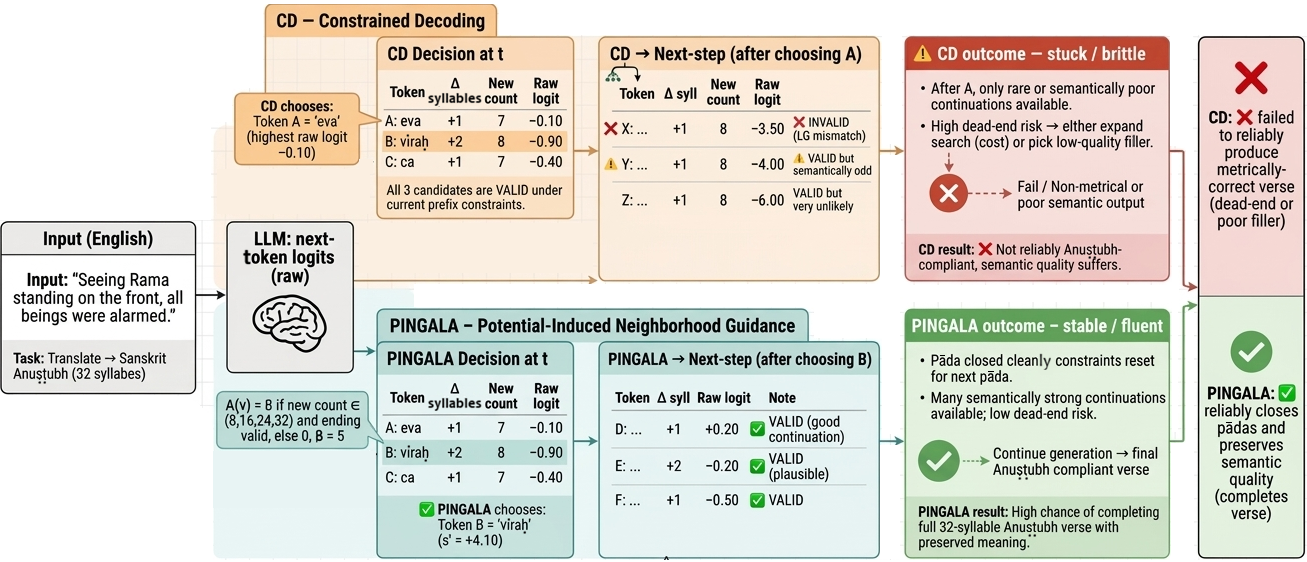}
    \caption{An infographic comparison of working of Pingala vs CD from \chandomitra.}
    \label{fig:placeholder}
\end{figure*}
\subsection{Pingala} 
\label{sec:math_formulation}
In this section, we define a new decoding strategy, beginning with general definition of variables in \autoref{sec:math_defn}, and subsequently mathematically formulate the decoding strategy from \autoref{sec:reward} to \autoref{sec:contrained-reward}.
\subsubsection{General Definitions}
\label{sec:math_defn}
\(x\) denotes the source (English) sentence, \(\mathcal{M}\) - the fine-tuned language model,  \(y=(y_1,\dots,y_T)\) a candidate Sanskrit token sequence, \(\mathcal{C}(\cdot)\in\{0,1\}\) - a prefix-aware metrical checker that returns 1 if and only if a Devanagari string satisfies the Anu\d{s}\d{t}ubh prefix constraints, \(P(z)\) - the laghu/guru pattern of string \(z\) (a sequence of \(L,G\) symbols) and \(|P(z)|\) its length in syllables .

\subsubsection{Prosody-aware shaping (p\={a}da weights)}
\label{sec:reward}
We define a function \(\Phi_{\text{meter}}(\cdot)\) over prefixes that captures p\={a}da/verse completion:
\begin{equation}
\begin{split}
    \Phi_{\text{meter}}(y_{1:t})
=\beta\sum_{j\in\{8,16,24,32\}}\mathbf{1}[|P(y_{1:t})|\!\ge\! j\land B_j] \\
\text{where  $B_j(y_{1:t})$ indicates that} \\ \text{the $j$-th metrical boundary is correct.}
\end{split}
\end{equation}
and use its discrete incremental difference as the per-step shaping weight:
\begin{equation}
\label{eq:step_reward}
\Delta(v; y_{1:t-1}) \;=\; \Phi_{\text{meter}}(y_{1:t-1}\Vert v) \;-\; \Phi_{\text{meter}}(y_{1:t-1})
\end{equation}
Equivalently, \(\Delta\) is non-zero only when the appended token completes a p\={a}da (or the whole verse) ; we set \(\beta>0\) (in this work, \(\beta=5\)) so that the additive adjustment multiplies the candidate's unnormalized selection weight by \(e^\beta\), strongly biasing search toward structurally stable prefixes.

\subsubsection{Constrained Logits Operator}
\label{sec:contraint-logit-op}
During decoding, the model produces raw logits \(s_t(v)\) for each vocabulary token \(v\) . We apply a constraint-aware transform \(\Phi\) to obtain modified logits \(s'_t(v)\) . This operator implements hard pruning plus additive shaping:
\begin{multline}
\label{eq:logits_transform}
s'_t(v) = -\infty
\quad \text{if } \mathcal{C}(\text{decode}(y_{1:t-1}\Vert v)) = 0;\\
s'_t(v) = s_t(v) + \Delta(v; y_{1:t-1})
\quad \text{otherwise.}
\end{multline}

where \(\Delta(v; y_{1:t-1})\) defined by \autoref{eq:step_reward} which is applied when a structural milestone is detected . Invalid candidates are set to \(-\infty\) and only the top-\(k\) tokens are scanned per step for computational efficiency .

\subsubsection{Constrained, Re-Weighted Decoding}
\label{sec:contrained-reward}
The decoding procedure approximates a constrained optimization that jointly maximizes model likelihood and metrical weight over the set of valid sequences \(\mathcal{Y}_{\text{valid}}=\{y:\mathcal{C}(y)=1\}\) . The practical objective being optimized by beam/lattice search with the logits transform (\autoref{eq:logits_transform}) is:
\begin{equation}
\label{eq:final_objective}
y^\star \;=\; \arg\max_{y \in \mathcal{Y}_{\text{valid}}} \; \Bigg( \log P_{\mathcal{M}}(y\mid x) \;+\; \lambda \cdot R_{\text{meter}}(y) \Bigg)
\end{equation}
where
\begin{equation}
R_{\text{meter}}(y) \;=\; \sum_{t=1}^{|y|} \Delta(y_t; y_{<t}).
\end{equation}
In practice, \(\lambda\) is absorbed into \(\beta\) (the additive constant) and the combined neural + symbolic score is realized stepwise through \autoref{eq:logits_transform} and \autoref{eq:step_reward}. EOS behavior is enforced by forcing \(s'_t(\text{EOS})\) large (and others \(-\infty\)) when \(|P(y_{1:t})|=32\), ensuring termination exactly at 32 syllables .


\subsection{Using SLP1 instead of Devanagari}
\label{sec:method:slp1}

We adopt SLP1 transliteration as a deliberate representation choice for Sanskrit inputs. The decision is motivated by several practical and modeling considerations grounded in prior work on tokenization, romanization, and Sanskrit processing. First, transliterating Devanagari into a compact ASCII phonemic form reduces tokenization mismatch with predominantly Latin-script pretrained vocabularies and therefore limits subword fragmentation and OOV effects that hinder efficient adaptation. \citep{limisiewicz2023tokenization,purkayastha2023romanization} Second, SLP1 provides a deterministic, normalization-stable mapping that avoids Unicode variation and normalization pitfalls common in Devanagari processing, improving robustness across preprocessing toolchains. \citep{adiga2021asr} Third, because SLP1 is phonemically oriented, it exposes sound-level regularities (e.g., vowel/consonant patterns and sandhi-related alternations) in a form that models can more readily learn from character- and subword-level statistics, which is important for meter- and phonology-sensitive tasks. \citep{nehrdich2024byt5sanskrit} Finally, an ASCII transliteration simplifies integration with existing tooling (tokenizers, prompt templates, and evaluation pipelines) and enables straightforward ablation and augmentation experiments without altering model internals. Together, these factors make SLP1 a low-cost, high-compatibility representation choice for downstream instruction tuning and metric-aware decoding in Sanskrit poetry modeling.

\section{Experimental Setup}
\label{sec:pipeline}

\begin{enumerate}[noitemsep, leftmargin=*]
  \item \textbf{Implementation of \autoref{sec:contrained-reward}}: During generation we employ a custom \texttt{LogitsProcessor} that performs three critical operations at each decode step: (a) prunes any token that immediately violates the Anu\d{s}\d{t}ubh prefix constraints using regex-based pattern matching (b) suppress EOS token until the verse reaches exactly 32 syllables, and (c) adds sparse additive weight (\(\beta=5.0\)) to tokens that complete metrical milestones (p\={a}da boundaries at syllables 8, 16, 24, or full verse at 32) . The processor is implemented in \autoref{alg:constrained_decoding_compact}. Further explanations are available in \autoref{sec:logits_processing}.
  \item \textbf{Ranking and fallback selection.} We generate multiple candidate sequences using beam search (beam size = 25, return 25 sequences) and apply post-filtering using a strict final checker . If one or more perfect 32-syllable verses exist, we select the top-ranked candidate . Otherwise, we employ a semantic fallback heuristic that prefers outputs with syllable counts closest to 32 and higher model likelihood scores. This staged decision ensures the system prioritizes perfect metric satisfaction while maintaining graceful degradation. Further explanation are available in \autoref{sec:beam_search}, \ref{sec:final_selection}.
  \item \textbf{SLP1 as transliteration scheme instead of Devanagri} during instrcution finetuning as well as while performing \texttt{Pingala}, our proposed deocding strategy, as described in \autoref{sec:exp_slp1} 
\end{enumerate}

\section{Results}
\label{sec:results}
Table~\ref{tab:main_result} summarizes syntactic and semantic evaluation on the \chandomitra ~test set consisting of 1421 English Prose - Sanskrit Poetry parallel data, for three settings: FT (fine-tuning on the Chandomitra training split), CD (constrained decoding as used in \chandomitra), and \texttt{Pingala} (our custom-constrained decoding). For syntax we report the fraction of outputs that are fully conformant to the Anu\d{s}\d{t}ubh meter (``Full \%'') and partially conformant (``Partial \%''); semantic fidelity is reported as automatic similarity (``Sim'') to the reference.

\noindent Key findings:
\begin{itemize}[noitemsep, leftmargin=*]
  \item Constrained decoding (CD/\texttt{Pingala}) substantially increases syntactic conformity but can interact with semantic fidelity depending on model adaptation.
  \item NLLB-dist-1.3B: applying CD improves syntax (Full \% = 82.97) while retaining good semantics (Sim = 70.75). Fine-tuning + \texttt{Pingala} yields perfect syntax (Full \% = 100) with strong semantics (Sim = 78.89).
  \item Phi-4-14B: \texttt{Pingala}-only runs can achieve very high semantic scores (Sim = 92.12) but with low syntactic coverage; FT + \texttt{Pingala} produces a balanced outcome (Full \% = 65.66, Sim = 76.75); using SLP1 as transliteration scheme for Sanskrit in FT+\texttt{Pingala} boosts the performance(Full \% = 95.92, Sim = 76.81) for metrical adherence while slightly increasing semantic similarity.
  \item Across architectures, the best trade-off is FT followed by \texttt{Pingala}: task-specific adaptation plus targeted decoding achieves near-perfect Anu\d{s}\d{t}ubh conformity with competitive semantic similarity.
\end{itemize}
\paragraph{Discussion.} Constrained decoding is an effective mechanism for enforcing meter, but it is most successful when the model has been exposed to the target poetic distribution (via FT). Practically, a two-stage recipe - (1) fine-tune on the target corpus, then (2) apply targeted constrained decoding -- yields metrically valid outputs while preserving semantic fidelity. A few samples of generated Sanskrit Poetry is available at \autoref{tab:data:sample-gen}.

\begin{table}[ht]
    \centering
    \resizebox{0.482\textwidth}{!}{
    \begin{tabular}{lcccccc}
        \toprule
        \textbf{Model} & \textbf{FT} & \textbf{CD} & \textbf{Pingala} & \textbf{Full \%} & \textbf{Partial \%} & \textbf{Sim} \\
        \midrule
        \multicolumn{7}{c}{\textbf{Raw Translation Models}} \\
        \midrule
        NLLB-dist-1.3B$^{*}$ & \cross & \tick & \cross & 82.97 & 90 & 70.75 \\
        NLLB-dist-1.3B & \cross & \cross & \tick & 0.14 & 0.25 & 35.62 \\
        NLLB-dist-1.3B$^{*}$ & \tick & \tick & \cross & \underline{99.09} & \underline{99.1} & 67.99 \\
        NLLB-dist-1.3B & \tick & \tick & \cross & 98.8 & 99.09 & 69.01 \\
        NLLB-dist-1.3B & \tick & \cross & \tick & \textbf{100} & \textbf{100} & 78.89 \\
        \midrule
        \multicolumn{7}{c}{\textbf{Instruction Following Models}} \\
        \midrule
        Phi-4-14B & \cross & \cross & \tick & 11.75 & 23.85 & \textbf{92.12} \\
        Phi-4-14B & \tick & \cross & \cross & 57.42 & 75.01 & 67.29 \\
        Phi-4-14B & \tick & \tick & \cross & 2.53 &	21.32 &	\underline{83.89} \\
        Phi-4-14B & \tick & \cross & \tick & 65.66 & 77.78 & 76.75 \\
        Phi-4-14B(SLP1) & \tick & \cross & \cross & 31.18 & 65.73 & 81.64 \\
        Phi-4-14B(SLP1) & \tick & \cross & \tick & 95.92 & 98.31 & 76.81 \\
        \bottomrule
    \end{tabular}
    }
    \caption{Evaluation on the Chandomitra test set. ``FT'' denotes fine-tuning on the \chandomitra training split; ``CD'' denotes standard constrained decoding as used in \chandomitra; ``\texttt{Pingala}'' denotes our custom-constrained decoding framework. ``Full \%'' is fraction of outputs fully conformant to Anu\d{s}\d{t}ubh; ``Partial \%'' is fraction partially conformant. ``Sim'' is automatic semantic similarity (higher is better). $^{*}$same settings as in \chandomitra}
    \label{tab:main_result}
\end{table}

\begin{table}[ht]
    \centering
    \resizebox{0.482\textwidth}{!}{
    \begin{tabular}{lccc}
        \toprule
        \textbf{Model} & \textbf{Full \%} & \textbf{Partial \%} & \textbf{Sim} \\
        \midrule
        NLLB-dist-1.3B & 100 & 100 & 69.76 \\
        Phi-4-14B & 61.54 & 74.23 & 74.62 \\
        Phi-4-14B(SLP1) & 92.31 & 95.85 & 67.87 \\
        \bottomrule
    \end{tabular}
    }
    \caption{Evaluation on the OOD set of \chandomitra. For the models that has been fine-tuned on training split of \chandomitra, and using Pingala. ``Full \%'' is fraction of outputs fully conformant to Anu\d{s}\d{t}ubh; ``Partial \%'' is fraction partially conformant. ``Sim'' is automatic semantic similarity (higher is better).}
    \label{tab:ood_result}
\end{table}

\section{Error Analysis}

\begin{table}[ht]
\centering
\small
\setlength{\tabcolsep}{8pt}
\resizebox{0.482\textwidth}{!}{
\begin{tabular}{lrrr}
\toprule
\textbf{Quality} & \textbf{NLLB} & \textbf{Phi-4} & \textbf{Phi-4(SLP1)} \\
\midrule
Excellent ($\ge0.8$) & 970  & 912 & 960 \\
Good (0.6--0.8) & 144  & 166 & 166 \\
Fair (0.4--0.6) & 95  & 89 & 111 \\
Poor (0.2--0.4) & 72 & 93 & 54 \\
Very poor ($<$0.2) & 140  & 161 & 130 \\
\bottomrule
\end{tabular}
}
\caption{Semantic similarity distribution for NLLB, Phi-4, and Phi-4(SLP1) outputs. Dataset size: $N=1421$.}
\label{tab:sem-error-analysis-box}
\end{table}

\begin{table}[ht]
\centering
\small
\setlength{\tabcolsep}{8pt}
\resizebox{0.482\textwidth}{!}{
\begin{tabular}{lrrr}
\toprule
\textbf{Syllable Deviation Category} & \textbf{NLLB} & \textbf{Phi-4} &  \textbf{Phi-4(SLP1)} \\
\midrule
Perfect (32) & 1421 & 1105 & 1397 \\
Near-perfect (28--36) & 0 & 206 & 23 \\
Moderate (24--40) & 0 & 104 & 0 \\
Poor & 0 & 6 & 1 \\
\bottomrule
\end{tabular}
}
\caption{Syntactic (Anu\d{s}\d{t}ubh meter) conformance analysis. Dataset size: $N=1421$.}
\label{tab:syntactic-error-analysis}
\end{table}

As shown in Table~\ref{tab:sem-error-analysis-box}, all three systems produce semantically excellent translations for the majority of samples, with NLLB and Phi-4(SLP1) leading. The low-scoring cases primarily stem from constrained decoding forcing lexical substitutions to satisfy metrical requirements, poor sub-word tokenization of rare Sanskrit forms causing hallucinations, and automatic metrics penalizing valid paraphrases that preserve metre but differ from the reference.

Table~\ref{tab:syntactic-error-analysis} reveals a complementary trade-off: NLLB's constrained decoding guarantees perfect metrical conformance but occasionally at the cost of semantic fidelity, while unconstrained Phi-4 preserves meaning more freely but frequently violates the 32-syllable structure. Phi-4(SLP1) strikes the best balance between these extremes, achieving near-perfect metrical conformance while maintaining competitive semantic similarity, making it the most effective configuration overall.

\section{Human Evaluation}
\begin{table}[ht]
    \centering
    \resizebox{0.482\textwidth}{!}{
    \begin{tabular}{ccc}
    \toprule
         \textbf{Annotator}& \textbf{Semantic Coherence}\\
         \midrule
         A1& 3.08\\
         A2& 2.32\\
         A3&2.5\\
         A4&1.96\\
         A5&2.28\\
         \midrule
         Krippendorff's alpha& 0.059\\
         \bottomrule
    \end{tabular}
    }
    \caption{Human Evaluation Results}
    \label{tab:placeholder}
\end{table}
We performed a focused human evaluation on a random sample of 50 test examples, obtained from the best performing model i.e. \texttt{NLLB-dist-1.3B}, independently rated by five expert reviewers. Each reviewer judged the same 50 outputs on \emph{semantic coherence} (degree of faithfulness to the input, 1–5 Likert). Reviewer backgrounds were diverse: Annotator 1 holds a Master's in Sanskrit literature and is engaged in pre-doctoral research; Annotator 2 is a practising poet and avadhani (Sanskrit and Telugu) who is also completing a technical Master's, Annotator 3 and 4 are trained in traditional Sanskrit pedagogy (Gurukulam background) with Master’s degrees in Sanskrit literature and Annotator 5 is an active Sanskrit scholar specializing in translation. The Inter-annotator agreement score $\alpha$ = 0.0585 which is low. 
For the generated poetry, majority of the reviewers agree on translation adequacy while disagreement stems from morpho-syntactic irregularities.

\section{Ablation}
\label{sec:ablation_beta}

We analyze the effect of the constraint weight \(\beta\), which controls the strength of pāda-completion weights during decoding, for the \texttt{NLLB-dist-1.3B} model. Intuitively, \(\beta\) governs the trade-off between enforcing metrical structure and preserving semantic fidelity. Larger values more aggressively bias the decoder toward p\={a}da-closing tokens, while smaller values allow the language model’s semantic prior to dominate.

\begin{table}[ht]
    \centering
    \resizebox{0.35\textwidth}{!}{
    \begin{tabular}{ccc}
    \toprule
    \textbf{$\beta$} & \textbf{Syntactic (\%)} & \textbf{Semantic (\%)} \\
    \midrule
    0  & 92.61  & 64.96 \\
    1  & 100.00 & 77.06 \\
    3  & 100.00 & 77.56 \\
    5  & 100.00 & \textbf{78.89} \\
    7  & 99.37  & 73.84 \\
    10 & 92.75  & 61.42 \\
    \bottomrule
    \end{tabular}
    }
    \caption{Ablation results for constraint weight \(\beta\).}
    \label{tab:ablation_beta}
\end{table}

\noindent
As shown in Table~\ref{tab:ablation_beta}, setting \(\beta=0\) (no weight) yields imperfect Anu\d{s}\d{t}ubh compliance and low semantic scores, indicating that hard constraint filtering alone is insufficient. Introducing a small weight (\(\beta=1\)–3) immediately restores full syntactic correctness while significantly improving semantic similarity. The best overall performance is achieved at \(\beta=5\), which maintains 100\% metrical compliance while maximizing semantic quality. For larger values (\(\beta \ge 7\)), semantic scores degrade sharply as decoding becomes overly dominated by structural weights, leading to metrically correct but semantically distorted outputs.

\begin{figure}[ht]
    \centering
    \includegraphics[width=0.48\textwidth]{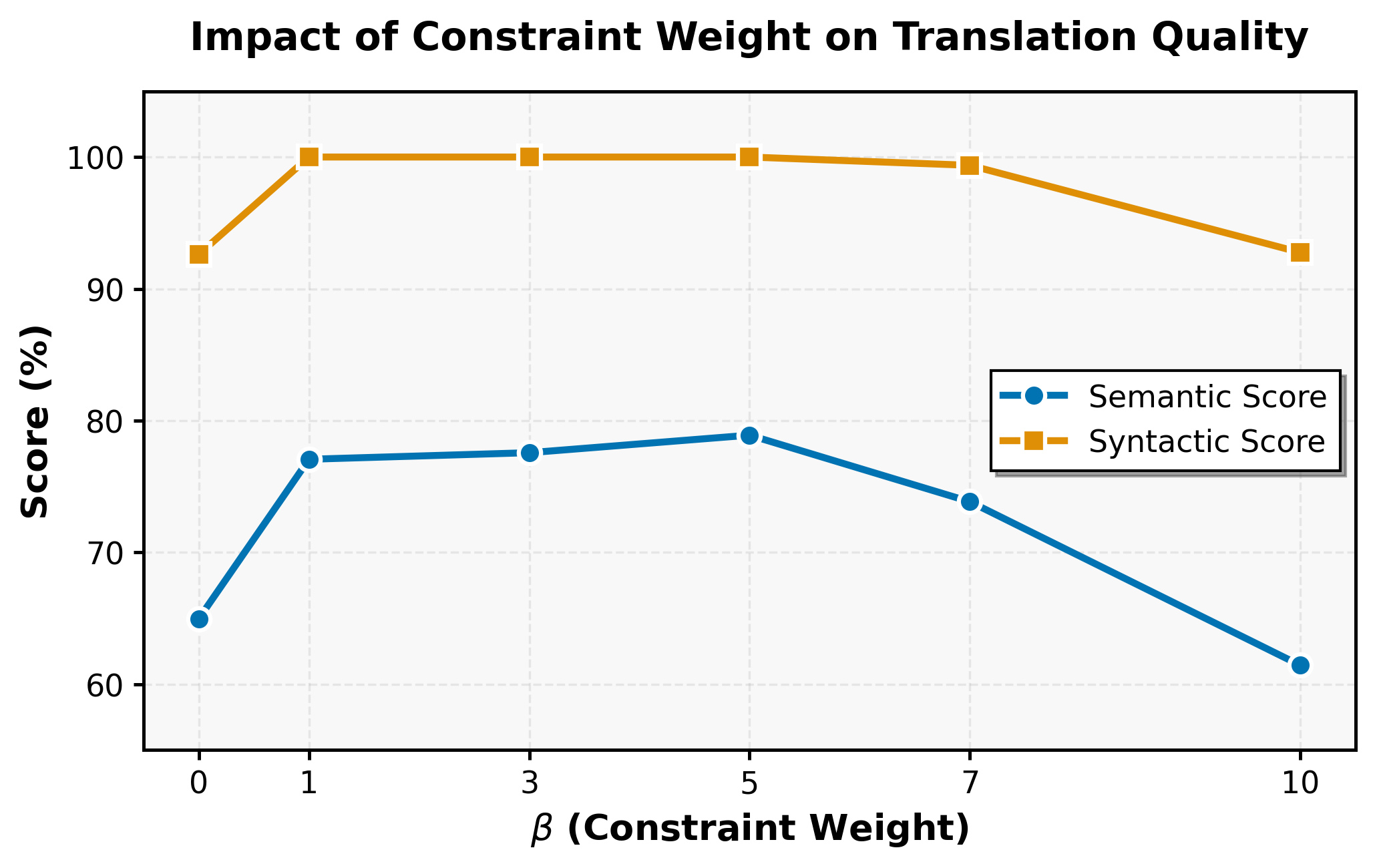}
    \caption{Impact of constraint weight \(\beta\) on syntactic and semantic quality. Semantic quality peaks at \(\beta=5\), illustrating the optimal balance between structural enforcement and meaning preservation.}
    \label{fig:beta_ablation}
\end{figure}

\noindent
Based on this ablation, we fix \(\beta=5\) for all experiments. This value consistently provides the best balance between structural stability and semantic fidelity across both NLLB and Phi-4 models.

\section{Conclusion}
In this work, we addressed the dual challenge of adhering to strict prosodic constraints while maintaining semantic coherence in Sanskrit poetry generation. We introduced Pingala, a model-agnostic, prosody-aware decoding framework applied seamlessly at inference time. By leveraging metrical milestones to strategically bias token selection, Pingala ensures adherence to the mandatory eight-syllable structure of the Anu\d{s}\d{t}ubh meter without requiring expensive retraining. Our findings highlight that combining task-specific fine-tuning with our targeted decoding strategy yields near-perfect metrical conformity alongside strong semantic fidelity. Crucially, we demonstrated that input representation matters: employing the phonemically aware SLP1 transliteration scheme dramatically amplifies the effectiveness of instruction fine-tuning for phonetically constrained tasks. To accurately measure these semantic improvements, we contributed a reference-free cross-encoder evaluation framework that demonstrates significantly higher correlation with expert human judgments than traditional bi-encoder approaches. \\
Looking forward, this research opens several exciting pathways. While Pingala successfully models the rigid Anu\d{s}\d{t}ubh meter, adapting these constraint-aware mechanisms to more complex, variable-length meters remains a vital next step. Furthermore, the low inter-annotator agreement observed during our semantic coherence evaluations underscores the need for larger rater pools and tighter rubrics to stabilize human judgments in low-resource computational philology. Beyond Sanskrit, extending these prosody-aware frameworks to other morphologically rich poetic traditions, such as Dravidian languages, presents a compelling opportunity to preserve their unique structural aesthetics. Finally, integrating multi-agent LLM systems to iteratively debate and refine these intricate metrical and semantic constraints could push the boundaries of computationally generated verse even further.
\section{Limitations}

Our work has the following limitations:

\begin{itemize}
    \item \textbf{Limited model scale:} Due to computational constraints, we use Phi-4 in an 8-bit quantized configuration and have not evaluated Pingala on larger language models (>10B parameters at full precision), which may exhibit different decoding dynamics. Extending our evaluation to such models remains important but was infeasible given our compute budget.

    \item \textbf{Small evaluator pool:} The human evaluation was conducted by only two expert annotators, as reliable assessment of Sanskrit poetry requires dual expertise in classical prosody and poetic composition, inherently limiting the pool of qualified evaluators. The low inter-annotator agreement (Cohen's $\kappa$ = 0.069) underscores that a larger expert panel would be desirable for more stable human judgments.

    \item \textbf{SLP1 not tested with NLLB:} Although we demonstrate substantial gains from SLP1 transliteration with Phi-4, we were unable to apply SLP1 to NLLB, as it was pre-trained exclusively on Devanagari-script data and continued pre-training on SLP1-encoded corpora was not feasible within our computational budget. The generalizability of SLP1-based improvements across model families thus remains an open question.
\end{itemize}

\section{Acknowledgement}
We gratefully acknowledge Rugved Deshpande, Senior Language Editor in the Department of Sanskrit Studies, University of Hyderabad, for his invaluable contributions to the human evaluation. This work was supported by the Google Gemma Academic Program Award and in part by the National Language Translation Mission (NLTM): Bhashini project of the Government of India.

\bibliography{new}

@misc{oord2019representationlearningcontrastivepredictive,
      title={Representation Learning with Contrastive Predictive Coding}, 
      author={Aaron van den Oord and Yazhe Li and Oriol Vinyals},
      year={2019},
      eprint={1807.03748},
      archivePrefix={arXiv},
      primaryClass={cs.LG},
      url={https://arxiv.org/abs/1807.03748}, 
}

@inproceedings{gao-etal-2021-simcse,
    title = "{S}im{CSE}: Simple Contrastive Learning of Sentence Embeddings",
    author = "Gao, Tianyu  and
      Yao, Xingcheng  and
      Chen, Danqi",
    editor = "Moens, Marie-Francine  and
      Huang, Xuanjing  and
      Specia, Lucia  and
      Yih, Scott Wen-tau",
    booktitle = "Proceedings of the 2021 Conference on Empirical Methods in Natural Language Processing",
    month = nov,
    year = "2021",
    address = "Online and Punta Cana, Dominican Republic",
    publisher = "Association for Computational Linguistics",
    url = "https://aclanthology.org/2021.emnlp-main.552/",
    doi = "10.18653/v1/2021.emnlp-main.552",
    pages = "6894--6910"
}

@misc{adi2017finegrainedanalysissentenceembeddings,
      title={Fine-grained Analysis of Sentence Embeddings Using Auxiliary Prediction Tasks}, 
      author={Yossi Adi and Einat Kermany and Yonatan Belinkov and Ofer Lavi and Yoav Goldberg},
      year={2017},
      eprint={1608.04207},
      archivePrefix={arXiv},
      primaryClass={cs.CL},
      url={https://arxiv.org/abs/1608.04207}, 
}

@inproceedings{reimers-gurevych-2020-making,
    title = "Making Monolingual Sentence Embeddings Multilingual using Knowledge Distillation",
    author = "Reimers, Nils  and
      Gurevych, Iryna",
    editor = "Webber, Bonnie  and
      Cohn, Trevor  and
      He, Yulan  and
      Liu, Yang",
    booktitle = "Proceedings of the 2020 Conference on Empirical Methods in Natural Language Processing (EMNLP)",
    month = nov,
    year = "2020",
    address = "Online",
    publisher = "Association for Computational Linguistics",
    url = "https://aclanthology.org/2020.emnlp-main.365/",
    doi = "10.18653/v1/2020.emnlp-main.365",
    pages = "4512--4525"
}

@misc{ramos2025finegrainedrewardoptimizationmachine,
      title={Fine-Grained Reward Optimization for Machine Translation using Error Severity Mappings}, 
      author={Miguel Moura Ramos and Tomás Almeida and Daniel Vareta and Filipe Azevedo and Sweta Agrawal and Patrick Fernandes and André F. T. Martins},
      year={2025},
      eprint={2411.05986},
      archivePrefix={arXiv},
      primaryClass={cs.CL},
      url={https://arxiv.org/abs/2411.05986}, 
}

@inproceedings{ranasinghe-etal-2020-transquest,
    title = "{T}rans{Q}uest: Translation Quality Estimation with Cross-lingual Transformers",
    author = "Ranasinghe, Tharindu  and
      Orasan, Constantin  and
      Mitkov, Ruslan",
    editor = "Scott, Donia  and
      Bel, Nuria  and
      Zong, Chengqing",
    booktitle = "Proceedings of the 28th International Conference on Computational Linguistics",
    month = dec,
    year = "2020",
    address = "Barcelona, Spain (Online)",
    publisher = "International Committee on Computational Linguistics",
    url = "https://aclanthology.org/2020.coling-main.445/",
    doi = "10.18653/v1/2020.coling-main.445",
    pages = "5070--5081"
}

@inproceedings{skrutable,
    title = "Skrutable: Another Step Toward Effective {S}anskrit Meter Identification",
    author = "Neill, Tyler",
    editor = "Kulkarni, Amba  and
      Hellwig, Oliver",
    booktitle = "Proceedings of the Computational {S}anskrit {\&} Digital Humanities: Selected papers presented at the 18th World {S}anskrit Conference",
    month = jan,
    year = "2023",
    address = "Canberra, Australia (Online mode)",
    publisher = "Association for Computational Linguistics",
    url = "https://aclanthology.org/2023.wsc-csdh.7/",
    pages = "97--112"
}

@inproceedings{chi-etal-2021-infoxlm,
  title = "{I}nfo{XLM}: An Information-Theoretic Framework for Cross-Lingual Language Model Pre-Training",
  author={Chi, Zewen and Dong, Li and Wei, Furu and Yang, Nan and Singhal, Saksham and Wang, Wenhui and Song, Xia and Mao, Xian-Ling and Huang, Heyan and Zhou, Ming},
  booktitle = "Proceedings of the 2021 Conference of the North American Chapter of the Association for Computational Linguistics: Human Language Technologies",
  month = jun,
  year = "2021",
  address = "Online",
  publisher = "Association for Computational Linguistics",
  url = "https://aclanthology.org/2021.naacl-main.280",
  doi = "10.18653/v1/2021.naacl-main.280",
  pages = "3576--3588",}

@inproceedings{jagadeeshan-etal-2026-chandomitra,
    title = "Chandomitra: Towards Generating Structured {S}anskrit Poetry from Natural Language Inputs",
    author = "Jagadeeshan, Manoj Balaji  and
      Bhatia, Samarth  and
      Ray, Pretam  and
      Surana, Harshul Raj  and
      P, Akhil Rajeev  and
      Mishra, Priya  and
      Kulkarni, Annarao  and
      Ramakrishnan, Ganesh  and
      Ap, Prathosh  and
      Goyal, Pawan",
    editor = "Demberg, Vera  and
      Inui, Kentaro  and
      Marquez, Llu{\'i}s",
    booktitle = "Proceedings of the 19th Conference of the {E}uropean Chapter of the {A}ssociation for {C}omputational {L}inguistics (Volume 1: Long Papers)",
    month = mar,
    year = "2026",
    address = "Rabat, Morocco",
    publisher = "Association for Computational Linguistics",
    url = "https://aclanthology.org/2026.eacl-long.24/",
    doi = "10.18653/v1/2026.eacl-long.24",
    pages = "518--534",
    ISBN = "979-8-89176-380-7"
}

@misc{bge-m3,
      title={BGE M3-Embedding: Multi-Lingual, Multi-Functionality, Multi-Granularity Text Embeddings Through Self-Knowledge Distillation}, 
      author={Jianlv Chen and Shitao Xiao and Peitian Zhang and Kun Luo and Defu Lian and Zheng Liu},
      year={2024},
      eprint={2402.03216},
      archivePrefix={arXiv},
      primaryClass={cs.CL}
}

@misc{nehrdich2026mitrasamgrahacomprehensiveclassicalsanskrit,
      title={Mitrasamgraha: A Comprehensive Classical Sanskrit Machine Translation Dataset}, 
      author={Sebastian Nehrdich and David Allport and Sven Sellmer and Jivnesh Sandhan and Manoj Balaji Jagadeeshan and Pawan Goyal and Sujeet Kumar and Kurt Keutzer},
      year={2026},
      eprint={2601.07314},
      archivePrefix={arXiv},
      primaryClass={cs.CL},
      url={https://arxiv.org/abs/2601.07314}, 
}

@inproceedings{reimers-gurevych-2019-sentence,
    title = "Sentence-{BERT}: Sentence Embeddings using {S}iamese {BERT}-Networks",
    author = "Reimers, Nils  and
      Gurevych, Iryna",
    editor = "Inui, Kentaro  and
      Jiang, Jing  and
      Ng, Vincent  and
      Wan, Xiaojun",
    booktitle = "Proceedings of the 2019 Conference on Empirical Methods in Natural Language Processing and the 9th International Joint Conference on Natural Language Processing (EMNLP-IJCNLP)",
    month = nov,
    year = "2019",
    address = "Hong Kong, China",
    publisher = "Association for Computational Linguistics",
    url = "https://aclanthology.org/D19-1410/",
    doi = "10.18653/v1/D19-1410",
    pages = "3982--3992"
}

@inproceedings{fomicheva-etal-2022-translation,
    title = "Translation Error Detection as Rationale Extraction",
    author = "Fomicheva, Marina  and
      Specia, Lucia  and
      Aletras, Nikolaos",
    editor = "Muresan, Smaranda  and
      Nakov, Preslav  and
      Villavicencio, Aline",
    booktitle = "Findings of the Association for Computational Linguistics: ACL 2022",
    month = may,
    year = "2022",
    address = "Dublin, Ireland",
    publisher = "Association for Computational Linguistics",
    url = "https://aclanthology.org/2022.findings-acl.327/",
    doi = "10.18653/v1/2022.findings-acl.327",
    pages = "4148--4159"
}

@inproceedings{rei-etal-2020-comet,
    title = "{COMET}: A Neural Framework for {MT} Evaluation",
    author = "Rei, Ricardo  and
      Stewart, Craig  and
      Farinha, Ana C  and
      Lavie, Alon",
    editor = "Webber, Bonnie  and
      Cohn, Trevor  and
      He, Yulan  and
      Liu, Yang",
    booktitle = "Proceedings of the 2020 Conference on Empirical Methods in Natural Language Processing (EMNLP)",
    month = nov,
    year = "2020",
    address = "Online",
    publisher = "Association for Computational Linguistics",
    url = "https://aclanthology.org/2020.emnlp-main.213/",
    doi = "10.18653/v1/2020.emnlp-main.213",
    pages = "2685--2702"
}

@misc{sinaga2025calibrationmeetsrealitymaking,
      title={Calibration Meets Reality: Making Machine Learning Predictions Trustworthy}, 
      author={Kristina P. Sinaga and Arjun S. Nair},
      year={2025},
      eprint={2509.23665},
      archivePrefix={arXiv},
      primaryClass={cs.LG},
      url={https://arxiv.org/abs/2509.23665}, 
}

@inproceedings{mesgar-etal-2023-devil,
    title = "The Devil is in the Details: On Models and Training Regimes for Few-Shot Intent Classification",
    author = "Mesgar, Mohsen  and
      Tran, Thy Thy  and
      Glava{\v{s}}, Goran  and
      Gurevych, Iryna",
    editor = "Vlachos, Andreas  and
      Augenstein, Isabelle",
    booktitle = "Proceedings of the 17th Conference of the European Chapter of the Association for Computational Linguistics",
    month = may,
    year = "2023",
    address = "Dubrovnik, Croatia",
    publisher = "Association for Computational Linguistics",
    url = "https://aclanthology.org/2023.eacl-main.135/",
    doi = "10.18653/v1/2023.eacl-main.135",
    pages = "1846--1857"
}

@misc{nllbteam2022languageleftbehindscaling,
      title={No Language Left Behind: Scaling Human-Centered Machine Translation}, 
      author={NLLB Team and Marta R. Costa-jussà and James Cross and Onur Çelebi and Maha Elbayad and Kenneth Heafield and Kevin Heffernan and Elahe Kalbassi and Janice Lam and Daniel Licht and Jean Maillard and Anna Sun and Skyler Wang and Guillaume Wenzek and Al Youngblood and Bapi Akula and Loic Barrault and Gabriel Mejia Gonzalez and Prangthip Hansanti and John Hoffman and Semarley Jarrett and Kaushik Ram Sadagopan and Dirk Rowe and Shannon Spruit and Chau Tran and Pierre Andrews and Necip Fazil Ayan and Shruti Bhosale and Sergey Edunov and Angela Fan and Cynthia Gao and Vedanuj Goswami and Francisco Guzmán and Philipp Koehn and Alexandre Mourachko and Christophe Ropers and Safiyyah Saleem and Holger Schwenk and Jeff Wang},
      year={2022},
      eprint={2207.04672},
      archivePrefix={arXiv},
      primaryClass={cs.CL},
      url={https://arxiv.org/abs/2207.04672}, 
}

@misc{abdin2024phi4technicalreport,
      title={Phi-4 Technical Report}, 
      author={Marah Abdin and Jyoti Aneja and Harkirat Behl and Sébastien Bubeck and Ronen Eldan and Suriya Gunasekar and Michael Harrison and Russell J. Hewett and Mojan Javaheripi and Piero Kauffmann and James R. Lee and Yin Tat Lee and Yuanzhi Li and Weishung Liu and Caio C. T. Mendes and Anh Nguyen and Eric Price and Gustavo de Rosa and Olli Saarikivi and Adil Salim and Shital Shah and Xin Wang and Rachel Ward and Yue Wu and Dingli Yu and Cyril Zhang and Yi Zhang},
      year={2024},
      eprint={2412.08905},
      archivePrefix={arXiv},
      primaryClass={cs.CL},
      url={https://arxiv.org/abs/2412.08905}, 
}

@inproceedings{limisiewicz2023tokenization,
  title     = {Tokenization Impacts Multilingual Language Modeling: Assessing Vocabulary Allocation and Overlap Across Languages},
  author    = {Tomasz Limisiewicz and Ji\v{r}\'{\i} Balhar and David Mare\v{c}ek},
  booktitle = {Findings of the Association for Computational Linguistics: ACL 2023},
  pages     = {5661--5681},
  year      = {2023},
  publisher = {Association for Computational Linguistics},
  url       = {https://aclanthology.org/2023.findings-acl.350}
}

@inproceedings{purkayastha2023romanization,
  title     = {Romanization-based Large-scale Adaptation of Multilingual Language Models},
  author    = {Sukannya Purkayastha and Sebastian Ruder and Jonas Pfeiffer and Iryna Gurevych and Ivan Vuli\'c},
  booktitle = {Findings of the Association for Computational Linguistics: EMNLP 2023},
  pages     = {7996--8005},
  year      = {2023},
  publisher = {Association for Computational Linguistics},
  url       = {https://aclanthology.org/2023.findings-emnlp.538}
}

@inproceedings{adiga2021asr,
  title     = {Automatic Speech Recognition in Sanskrit: A New Speech Corpus and Modelling Insights},
  author    = {Devaraja Adiga and Rishabh Kumar and Amrith Krishna and Preethi Jyothi and Ganesh Ramakrishnan and Pawan Goyal},
  booktitle = {Findings of the Association for Computational Linguistics: ACL-IJCNLP 2021},
  pages     = {5039--5050},
  year      = {2021},
  publisher = {Association for Computational Linguistics},
  url       = {https://aclanthology.org/2021.findings-acl.447}
}

@inproceedings{nehrdich2024byt5sanskrit,
  title     = {One Model is All You Need: ByT5-Sanskrit, a Unified Model for Sanskrit},
  author    = {Sebastian Nehrdich and Oliver Hellwig and Kurt Keutzer},
  booktitle = {Findings of the Association for Computational Linguistics: EMNLP 2024},
  pages     = {13742--13751},
  year      = {2024},
  publisher = {Association for Computational Linguistics},
  url       = {https://aclanthology.org/2024.findings-emnlp.805}
}

@inproceedings{
gao2018representation,
title={Representation Degeneration Problem in Training Natural Language Generation Models},
author={Jun Gao and Di He and Xu Tan and Tao Qin and Liwei Wang and Tieyan Liu},
booktitle={International Conference on Learning Representations},
year={2019},
url={https://openreview.net/forum?id=SkEYojRqtm},
}

@inproceedings{rajaee2022isotropy,
  title={An isotropy analysis in the multilingual BERT embedding space},
  author={Rajaee, Sara and Pilehvar, Mohammad Taher},
  booktitle={Findings of the Association for Computational Linguistics: ACL 2022},
  pages={1309--1316},
  year={2022}
  }

@inproceedings{
Humeau2020Poly-encoders:,
title={Poly-encoders: Architectures and Pre-training Strategies for Fast and Accurate Multi-sentence Scoring},
author={Samuel Humeau and Kurt Shuster and Marie-Anne Lachaux and Jason Weston},
booktitle={International Conference on Learning Representations},
year={2020},
url={https://openreview.net/forum?id=SkxgnnNFvH}
}

@article{rosa2022defense,
  title={In defense of cross-encoders for zero-shot retrieval},
  author={Rosa, Guilherme and Bonifacio, Luiz and Jeronymo, Vitor and Abonizio, Hugo and Fadaee, Marzieh and Lotufo, Roberto and Nogueira, Rodrigo},
  journal={arXiv preprint arXiv:2212.06121},
  year={2022}
}

@inproceedings{risch-etal-2021-semantic,
    title = "Semantic Answer Similarity for Evaluating Question Answering Models",
    author = {Risch, Julian  and
      M{\"o}ller, Timo  and
      Gutsch, Julian  and
      Pietsch, Malte},
    editor = "Fisch, Adam  and
      Talmor, Alon  and
      Chen, Danqi  and
      Choi, Eunsol  and
      Seo, Minjoon  and
      Lewis, Patrick  and
      Jia, Robin  and
      Min, Sewon",
    booktitle = "Proceedings of the 3rd Workshop on Machine Reading for Question Answering",
    month = nov,
    year = "2021",
    address = "Punta Cana, Dominican Republic",
    publisher = "Association for Computational Linguistics",
    url = "https://aclanthology.org/2021.mrqa-1.15/",
    doi = "10.18653/v1/2021.mrqa-1.15",
    pages = "149--157"
}

@inproceedings{sellam-etal-2020-bleurt,
    title = "{BLEURT}: Learning Robust Metrics for Text Generation",
    author = "Sellam, Thibault  and
      Das, Dipanjan  and
      Parikh, Ankur",
    editor = "Jurafsky, Dan  and
      Chai, Joyce  and
      Schluter, Natalie  and
      Tetreault, Joel",
    booktitle = "Proceedings of the 58th Annual Meeting of the Association for Computational Linguistics",
    month = jul,
    year = "2020",
    address = "Online",
    publisher = "Association for Computational Linguistics",
    url = "https://aclanthology.org/2020.acl-main.704/",
    doi = "10.18653/v1/2020.acl-main.704",
    pages = "7881--7892"
}

\appendix
\onecolumn

\section{Instruction Fine-tuning Prompt}
\label{sec:prompt}
\begin{figure}[h!]
\centering
\resizebox{!}{9cm}{
    \includegraphics{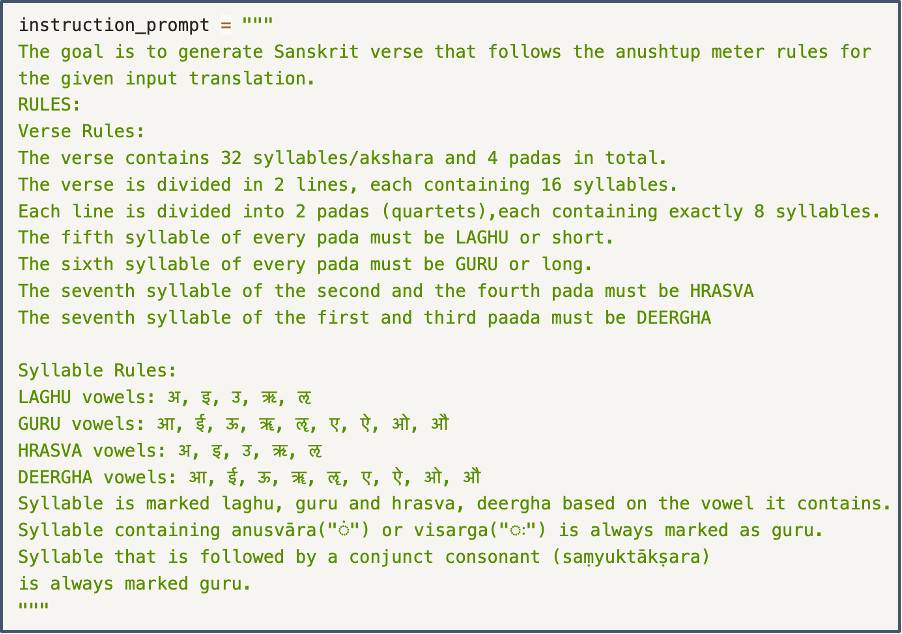}
    }
    \caption{Instruction fine-tuning Prompt}
    \label{image/IFT-prompt}
\end{figure}
\autoref{image/IFT-prompt} depicts the prompt used for instruction fine-tuning.

\section{Experimentation Details}
\subsection{Constrained, Prosody-Aware Shaping and Re-Weighted Decoding}
\label{sec:logits_processing}

Processing  
At generation step \(t\), the base model produces raw logits \(s_t(v)\) for each vocabulary token \(v\). We replace these by modified logits \(s'_t(v)\) using the combined operator described in Sections 4.12–4.14: prosody-aware shaping (Section 4.12), the constrained logits operator (Section 4.13) and the re-weighted decoding objective (Section 4.14).

The integrated operator is implemented efficiently by the following practical steps:
\begin{itemize}
  \item Decoding only the top-\(k\) candidates (default \(k=100\)) from the language model at every step.
  \item Scanning each prefix+candidate concatenation for syllable weights with a cached scanner to avoid redundant work.
  \item Applying pre-compiled regex-based partial-pattern checks for each \={p}āda position to detect milestone completions and partial matches.
  \item Masking tokens that violate hard structural constraints (setting their logits to \(-\infty\)) so they are never considered by the search.
  \item Adding an incremental shaping weights to candidates that complete desirable metrical milestones (implemented as an additive boost to rescored logits).
  \item Enforcing EOS/new-verse behavior explicitly when the verse-level structural target is reached (e.g., termination at the final syllable boundary).
  \item Adaptively expanding the search space (increasing \(k\)) and re-scoring when no valid candidates are found in the initial top-\(k\).
\end{itemize}

Implementation notes and heuristics:
\begin{itemize}
  \item Use of a cached scanner plus pre-compiled regexes keeps per-token checks lightweight and amenable to batching.
  \item Additive shaping is only applied when a structural milestone (p\=ada completion or similar) is detected, ensuring the language model's raw semantic score dominates unless structural signals are present.
  \item Hard masking is used for unequivocal structural violations; shaping is used for soft guidance where multiple continuations remain plausible.
  \item EOS and special-token handling are adjusted to guarantee exact verse termination when the full structural target is met.
\end{itemize}

\paragraph{Complexity analysis}: The per-step overhead introduced by constraint checking and shaping is proportional to the number of rescored candidates and the cost of scanning/regex checks. In practice the per-step complexity is \(O(k\cdot m)\) where \(k\) is the number of rescored tokens and \(m\) is the average scanning/matching cost. Limiting \(k\) (e.g., to 100) and using cached scans keeps latency reasonable even for long generations.

Practical behaviour and robustness:
\begin{itemize}
  \item The hard-mask + shaping hybrid preserves the base model's semantic ranking while proactively steering search toward structurally feasible prefixes, yielding higher rates of metrically valid completions.
  \item Adaptive expansion of the candidate set prevents dead-ends: when the initial top-\(k\) contains no valid continuations, the algorithm enlarges the search and retries, prioritizing feasibility without blind brute-force.
  \item Periodic cache clearing and careful token filtering reduce false positives from non-Devanagari tokens and control memory/latency trade-offs during batched generation.
\end{itemize}

\subsection{Beam Search and Hypothesis Management}
\label{sec:beam_search}

We do not commit greedily . Instead, we run a beam/lattice style search implemented via the model's native beam generation with the logits processor plugged in . The logits processor both prunes invalid branches and re-ranks candidates inside the beam according to the combined neural + re-weighting. This keeps multiple hypotheses alive and prevents premature, irreversible commitments to structurally fragile tokens - the principal failure mode of filter-only methods .

\textbf{Hyperparameters:} In our experiments we used beam size = 25, number of returned sequences = 25, length penalty = 1.0, and no-repeat-ngram size = 3 . These values were tuned empirically on a validation set and are configurable in the generation wrapper .

\subsection{Final Selection and Semantic Fallback}
\label{sec:final_selection}

From the generated set of returned candidates, we apply a strict final checker defined as \(\mathcal{C}(\cdot)\)(refer to \autoref{sec:math_defn}): any candidate that fails hard metrical rules is removed . If one or more perfect 32-syllable Anu\d{s}\d{t}ubh verses exist, we select the top one (first in returned beam order) . If none exist, we select a semantic fallback candidate using a heuristic ranking function:
\begin{equation}
\begin{split}
\text{score}_{\text{fallback}}(y) = -\alpha \cdot ||P(y)| - 32| \\ + \log P_{\mathcal{M}}(y\mid x) + \gamma \cdot \mathcal{C}_{\text{partial}}(y)
\end{split}
\end{equation}
where \(\alpha\) penalizes syllable count deviation, \(\gamma\) rewards partially valid prefixes, and \(\mathcal{C}_{\text{partial}}\) counts the number of correctly completed pādas . This staged decision ensures the system prioritizes perfect metric satisfaction but still returns a meaningful translation when exact meter cannot be formed .

\begin{algorithm}[ht]
\caption{Anu\d{s}\d{t}ubh Constrained Decoding}
\label{alg:constrained_decoding_compact}
\resizebox{0.87\linewidth}{!}{
\begin{minipage}{\linewidth}
\DontPrintSemicolon
\KwIn{Model $\mathcal{M}$, tokenizer $\mathcal{T}$, input $x$, beam $B$, top-$k$ $k$, weight $\beta$, $T_{\max}$}
\KwOut{Best verse $y^\star$ or semantic fallback}

$\mathcal{B}\leftarrow\{(\langle\mathrm{BOS}\rangle,0)\}$; $t\leftarrow 1$\;

\While{$\mathcal{B}$ has incomplete hyp. \textbf{and} $t\le T_{\max}$}{
  $\mathcal{B}_{\text{new}}\leftarrow\varnothing$\;
  \ForEach{ $(y_{1:t-1}, s_{1:t-1}) \in \mathcal{B}$ }{
    $s_t \leftarrow \mathcal{M}.\text{forward}(y_{1:t-1})$;\quad
    $\mathcal{V}_{\text{top}}\leftarrow\text{TopK}(s_t,k)$\;
    \ForEach{$v\in\mathcal{V}_{\text{top}}$}{
      $z\leftarrow\mathcal{T}.\text{decode}(y_{1:t-1}\Vert v)$;\quad
      $P\leftarrow\text{SyllableScanner}(z)$\;
      \If{ $\neg\mathcal{C}(z)$ }{ continue }                
      $\Delta\leftarrow \beta\cdot\mathbf{1}\{\,|P|\in\{8,16,24,32\}\wedge\text{PadaBoundaryValid}(z)\}$\;
      $s'\leftarrow s_t(v)+\Delta$\;
      \If{$|P|=32$}{
        \If{$v=\text{EOS}$}{$s'\leftarrow s'+100$} \Else{$s'\leftarrow -\infty$}
      }
      add $(y_{1:t-1}\Vert v,\; s_{1:t-1}+s')$ to $\mathcal{B}_{\text{new}}$\;
    }
  }
  $\mathcal{B}\leftarrow\text{TopBeam}(\mathcal{B}_{\text{new}},B)$;\quad $t\leftarrow t+1$\;
}

$\mathcal{Y}_{\text{complete}}\leftarrow\{y\in\mathcal{B}:\text{is\_complete}(y)\}$\;
$\mathcal{Y}_{\text{perfect}}\leftarrow\{y\in\mathcal{Y}_{\text{complete}}:|P(y)|=32\wedge\mathcal{C}(y)=1\}$\;

\uIf{$\mathcal{Y}_{\text{perfect}}\neq\varnothing$}{
  $y^\star\leftarrow\arg\max_{y\in\mathcal{Y}_{\text{perfect}}}P_{\mathcal{M}}(y\mid x)$\;
}
\Else{
  $y^\star\leftarrow\arg\max_{y\in\mathcal{Y}_{\text{complete}}}\text{score}_{\text{fallback}}(y)$\;
}

\Return{$y^\star$}\;
\end{minipage}
}
\end{algorithm}

\subsection{Top-k Validation and Performance Trade-offs}
\label{sec:topk_validation}

To keep inference tractable, we only check the top-\(k\) tokens per step (\(k=100\) by default) . This represents a trade-off: larger \(k\) reduces false negatives (missed valid choices) but increases the computational overhead of syllable scanning and regex validation . The logits processor includes adaptive logic to expand the checked set when the initial top-\(k\) yields no valid candidates .

\textbf{Adaptive expansion strategy:} If all top-\(k\) tokens are invalid, we double \(k\) (up to a maximum of 500) and rescan . This rarely occurs in practice after fine-tuning, as the model learns to place valid continuations higher in its distribution .

\subsection{Implementation details for using SLP1}
\label{sec:exp_slp1}
In addition to experiments conducted on Devanagari-script data, we include an experimental variant that operates entirely in the SLP1 transliteration space. To enable this, all Sanskrit text in the existing dataset is deterministically converted from Devanagari to SLP1 using the \texttt{indic-transliteration} Python library.\footnote{\url{https://github.com/indic-transliteration/indic-transliteration}} This conversion is applied uniformly to the training, validation, and evaluation splits to ensure representation consistency.

\noindent Following transliteration, we perform instruction fine-tuning of \texttt{Phi-4-14B} model using the SLP1-encoded Anu\d{s}\d{t}ubh samples. To avoid representation mismatch between instructions and model outputs, the initial system and task prompts are also modified such that all Sanskrit components originally written in Devanagari are transliterated into SLP1; this prevents the model from simultaneously conditioning on mixed-script representations, which could otherwise reintroduce tokenization inconsistencies during instruction following.

\noindent At inference time, we apply \texttt{Pingala}, our metric-aware constrained decoding method, directly on the SLP1 token stream. Since SLP1 preserves phonological structure in a normalized and tokenizer-friendly form, it integrates seamlessly with the constraint checks required for metrical validation during decoding. Additionally, we enforce two phonological constraints during beam search: first, a consonant cluster constraint that rejects any candidate token producing a saṃyuktākṣara (conjunct consonant) with more than three consecutive half-akṣaras (ardha-ak\d{s}aras), ensuring phonologically plausible output; and second, a hard syllable cap that assigns a score of $-\infty$ to any token whose addition would push the total syllable count beyond 32, thereby preventing out-of-bound generation and keeping all beam candidates within the valid anu\d{s}\d{t}ubh length.. Apart from the changes defined previously, all other experimental settings i.e. model architecture, optimization parameters, and decoding hyperparameters, are kept identical to the Devanagari-based setup, allowing for controlled comparison.

\textbf{Note:} We employ the SLP1 transliteration scheme only for experiments involving the Phi-4 model. We do not apply SLP1 to NLLB-dist-1.3B, as this model was pre-trained on Sanskrit data in the Devanagari script. For semantic evaluation, we consistently use Devanagari: in our empirical analysis, as seen in \autoref{tab:human_corrc_dev_vs_slp1}, the cross-encoder fine-tuned on Devanagari exhibits higher correlation with human judgments than its counterpart fine-tuned on SLP1. Accordingly, Devanagari is retained for all semantic similarity experiments. Even for the SLP1 generated poetry, we transliterate it to Devanagari and then evaluate semantic similarity. \autoref{sec:slp1-for-crossencoder} provides a detailed analysis comparing Devanagari and SLP1 as transliteration schemes for Sanskrit when fine-tuning a cross-encoder for semantic alignment with English inputs.


\begin{table}
    \centering
    \resizebox{\textwidth}{!}{
    \begin{tabular}{ll}
        \toprule
        \textbf{Model}& \textbf{Generated Sanskrit Poetry}\\
         \midrule
         NLLB-dist-1.3B& \devanagari{तस्याहं न सकामां कृत्स् संस्काराणां महायशाः ।दुःखोपजीवितायास्त्वत् कालयोगेन केनचिन् ॥}\\
         Phi-4-14B& \devanagari{अक्षमश्च मया कार्षी श्शोकपरा महायशाः । कृतं कर्म महत्कार्यं किं पुनर्जन्म जीवित ।।}\\
         Phi-4-14B(SLP1)& na ca tasya mahAtmAnaH kriyatAmitaraM kriyam . mama duzkftakAlasya jIvitA kiM Bavizyati ..\\
         \bottomrule
    \end{tabular}
    }
    \caption{Sample of generated Sanskrit poetry for the input "I could not even perform the last rites of that magnanimous one who died of grief on account of me. What use is my life with an unlucky birth?"}
    \label{tab:data:sample-gen}
\end{table}

\section{Appendix for Cross-Encoder as Semantic Evaluator}
\subsection{Background: Bi-Encoder semantic similarity and its limitations.}
Our initial translation-verification pipeline used a \emph{bi-encoder} architecture, similar to \chandomitra - the BGE-M3\footnote{\url{https://huggingface.co/BAAI/bge-m3}} \cite{bge-m3} model fine-tuned on \texttt{Mitrasa\d{m}graha} \cite{nehrdich2026mitrasamgrahacomprehensiveclassicalsanskrit}. In this formulation, the English source \(e\), and a candidate Sanskrit translation \(s\) are encoded independently by a shared encoder \(f_{\phi}\) into fixed-dimensional vectors:
\begin{align}
v_e &= f_{\phi}(e), \label{eq:vec_en} \\
v_s &= f_{\phi}(s). \label{eq:vec_sans}
\end{align}
Semantic similarity is then computed using cosine similarity:
\begin{equation}
\label{eq:cosine_sim}
\mathrm{sim}(e,s) \;=\; \cos(v_e, v_s)
\;=\;
\frac{v_e^\top v_s}{\|v_e\|\,\|v_s\|}.
\end{equation}
The similarity score is calculated using the linearized cosine similarity based
semantic similarity metric, similar to \chandomitra  work:
$$\text{score}_i = \frac{\pi - \cos^{-1}(\cos(\emb{x_i}, \emb{\hat{y}_i}))}{\pi}$$


Contrastive objectives (e.g., SimCSE~\cite{gao-etal-2021-simcse}, InfoNCE~\cite{oord2019representationlearningcontrastivepredictive}, triplet loss~\cite{reimers-gurevych-2020-making}) train bi-encoders to score aligned English–Sanskrit pairs higher than mismatches. While efficient and scalable, this paradigm depends on \emph{global} sentence embeddings, so the resulting similarity primarily measures coarse semantic proximity rather than the fine-grained token-level correspondences (word order, alignment, syntactic structure) needed for translation fidelity~\cite{adi2017finegrainedanalysissentenceembeddings,ramos2025finegrainedrewardoptimizationmachine}.

\paragraph{Limitations of bi-encoders for translation verification.} As reported in Table~\ref{tab:semantic-score-spread}, the average similarity gap between positives and hard negatives saturates at $\approx 22\%$, suggesting an intrinsic representational ceiling imposed by independent encoding and fixed-vector geometry. Table~\ref{tab:semantic-human} further exposes poor calibration: the bi-encoder’s mean similarity on ground-truth pairs is 73.46\% (human score: 5/5) but only slightly lower (69.05\%) on candidate outputs whose human ratings drop to 2.7/5. In short, the bi-encoder assigns similar numeric scores to qualitatively different translations, undermining its reliability for translation verification.
\subsection{Rationale --> Gradient-Level Explanation for using Class-Weighted binary cross-entropy}
\label{appendix:class-weighted}
Differentiating the loss with respect to the logit yields:
\[
\frac{\partial \mathcal{L}}{\partial g_{\theta}(x)}
=
\frac{1}{|\mathcal{D}|} w(y)\big(p_{\theta}(x)-y\big).
\]
For positive examples ($y=1$), gradient descent increases $g_{\theta}(x)$ and hence $p_{\theta}(x)$. For negative examples ($y=0$), gradient descent decreases $g_{\theta}(x)$ and hence $p_{\theta}(x)$. In expectation, this simultaneously raises $\mathbb{E}[p_{\theta}\mid y=1]$ and lowers $\mathbb{E}[p_{\theta}\mid y=0]$, directly increasing the probability gap.

\subsection{Motivation for the Cross-Encoder.}
To address the representational and calibration shortcomings of bi-encoders, we adopt a \emph{cross-encoder}, in our case - infoxlm\footnote{\url{https://huggingface.co/microsoft/infoxlm-large}} \cite{chi-etal-2021-infoxlm}. A cross-encoder takes the English and Sanskrit texts \emph{together} as input to a single Transformer stack so that full cross-attention between source and candidate is available. Formally, for a pair \((e,s)\) we feed the concatenated sequence (with appropriate separators) to a Transformer and compute a scalar score:
\begin{equation}
\label{eq:cross_enc_score}
\text{score}(e,s) \;=\; g_{\psi}\big(\, \mathrm{Transformer}\big([\texttt{CLS}\; e\; \texttt{[SEP]}\; s]\big) \,\big),
\end{equation}
where \(g_{\psi}(\cdot)\) is a small parameterized head (e.g., an MLP over the pooled representation) producing a calibrated likelihood that \(s\) is a correct translation of \(e\).

\noindent The cross-encoder’s direct token-level interaction confers several concrete advantages for translation verification:
\begin{itemize}[noitemsep]
  \item \textbf{Fine-grained alignment:} cross-attention enables the model to check lexical correspondences, agreement, and argument structure directly. \cite{reimers-gurevych-2019-sentence, ranasinghe-etal-2020-transquest}
  \item \textbf{Calibration:} as a discriminative scorer, the cross-encoder produces logits that are better separated between positive and negative pairs, which improves thresholding and ranking. \cite{rei-etal-2020-comet}
  \item \textbf{Error sensitivity:} local phrase-level errors (e.g., incorrect inflection or missing negation) produce pronounced changes in the cross-encoder score. \cite{fomicheva-etal-2022-translation}
\end{itemize}

\subsubsection{Why Cross-Encoder Outperforms Bi-Encoder}

Unlike the bi-encoder similarity score, \(\cos(v_e, v_s),\), which depends on independent global embeddings, the cross-encoder computes \(g_{\theta}(e,s)\), conditioned on full cross-attention between tokens. This allows the model to evaluate lexical alignment, syntactic agreement, and semantic role consistency directly. The resulting logit distributions exhibit larger separation and lower variance, producing a substantially higher probability gap after sigmoid normalization.

\subsection{Rationale on why the spread is high for Cross-encoder Model}
\label{appendix:spread-ratonale}
Logit Distribution Analysis: Let the class-conditional logits follow:
\[
Z_1 \sim \mathcal{N}(\mu_1, \sigma_1^2), \qquad
Z_0 \sim \mathcal{N}(\mu_0, \sigma_0^2).
\]
Using a logistic-normal approximation, the mean predicted probability is:
\[
\mathbb{E}[\sigma(Z)] \approx
\sigma\!\Big(\frac{\mu}{\sqrt{1 + c\sigma^2}}\Big),
\quad c=\tfrac{\pi^2}{3}.
\]
Thus,
\[
G \approx
\sigma\!\Big(\frac{\mu_1}{\sqrt{1+c\sigma_1^2}}\Big)
-
\sigma\!\Big(\frac{\mu_0}{\sqrt{1+c\sigma_0^2}}\Big).
\]
Curriculum hard-negative mining increases $\Delta\mu=\mu_1-\mu_0$ while reducing $\sigma_0^2$, yielding a strong nonlinear amplification of $G$.

\subsection{Qualitative Analysis of Output}
\label{appendix:qualitative-ce}
We performed a focused human evaluation on a random sample of \(N=200\) test examples. A single expert rater (a practising poet and \emph{avadhani} in Sanskrit and Telugu who is also completing a technical Master’s) scored each candidate translation for \emph{semantic fidelity} on a 1--5 Likert scale (1 = not faithful, 5 = fully faithful). We then report the correlation between the human scores and automatic similarity scores produced by the bi-encoder and cross-encoder verifiers, both before and after isotonic calibration \cite{sinaga2025calibrationmeetsrealitymaking}.

\section{Using SLP1 as Transliteration Scheme for Sanskrit}
\subsection{Instruction Fine-tuning Prompt for SLP1}
\label{sec:prompt-slp1}
\begin{figure}[h!]
\centering
\resizebox{!}{7cm}{
    \includegraphics{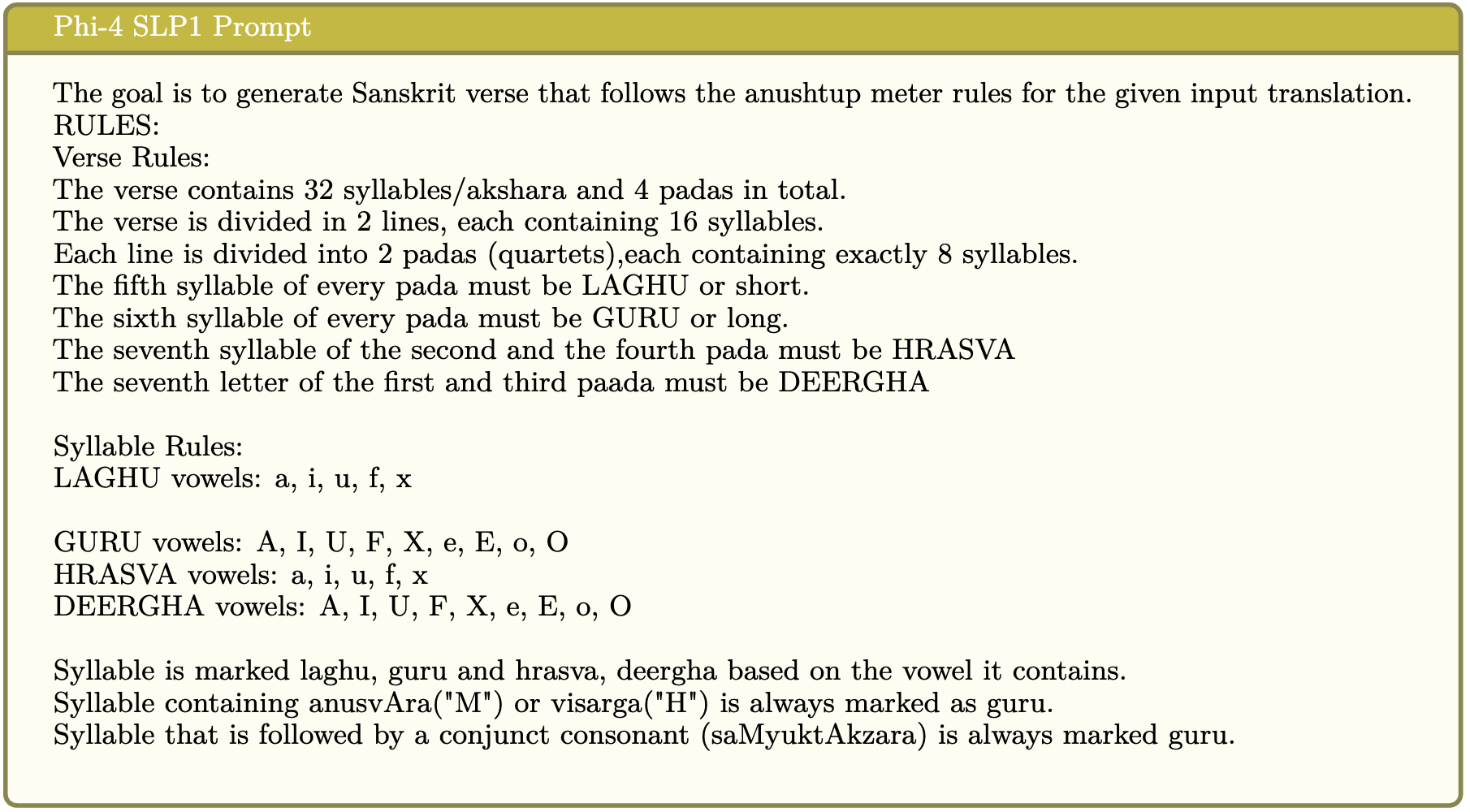}
    }
    \caption{Phi-4 SLP1 Instruction fine-tuning Prompt}
    \label{image/IFT-prompt-SLP1}
\end{figure}

\autoref{image/IFT-prompt-SLP1} shows the prompt used for instruction fine-tuning for generating Sanskrit Anu\d{s}\d{t}ubh poetry in SLP1 transliteration.

\subsection{Considering SLP1 for Sanskrit in Cross-Encoder}
\label{sec:slp1-for-crossencoder}
As described in \autoref{sec:method:slp1}, \autoref{sec:exp_slp1} and \autoref{sec:results}, adopting SLP1 for transliteration improves our method’s metrical conformity. This prompted the question: \textit{How does using SLP1 (rather than Devanagari) affect performance of the cross-encoder?} To answer this, we re-ran the experiment from \autoref{sec:semantic_similarity} on the Sanskrit data after converting it to SLP1. \autoref{tab:semantic-acc_dev_vs_slp1} reports mean accuracy for the two transliteration schemes; there is no substantial difference in overall accuracy. \autoref{tab:semantic-score-spreadc_dev_vs_slp1} compares the semantic-similarity score distributions: the spread values are similar, although the positive and negative(avg) scores are shifted by $\approx$5 units in the SLP1 condition relative to Devanagari. Finally, to decide which encoder to use in practice we computed correlations between each model’s scores and human ratings (\autoref{tab:human_corrc_dev_vs_slp1}). The cross-encoder trained on Devanagari shows higher correlation with human judgments than the model trained on SLP1. Taken together, these results indicate that while SLP1 yields measurable benefits for metrical conformity during the poetry generation task, it does not improve semantic agreement with human annotations - in our setup slightly degrades; therefore we retain the Devanagari-trained cross-encoder for semantic tasks.
\begin{table}
    \centering
    
    \begin{tabular}{cccc}
    \toprule
         \textbf{Approach}&  \textbf{Positive}&  \textbf{Negative(Avg.)}& \textbf{Spread}\\
    \midrule
        Devanagri & 90.02\% & 1.67\% & 88.33\%\\
        SLP1 & 95.81\% & 6.07\% & 89.74\%\\
    \bottomrule
    \end{tabular}
    
    \caption{Cross-encoder semantic similarity score distribution: Devagari vs SLP1 transliteration scheme}
    \label{tab:semantic-score-spreadc_dev_vs_slp1}
\end{table}

\begin{table}
    \centering
    \begin{tabular}{ccc}
    \toprule
    \multicolumn{3}{c}{\textbf{Mean Acc.}} \\
         \textbf{Batch Size}&  \textbf{Devanagari}& \textbf{SLP1}\\
         \midrule
         Full&  74.74\% & 77.07\%\\
         16&  84.71\% & 78.91\%\\
         4&  92.87\% & 90.07\%\\
         \bottomrule
    \end{tabular}
    \caption{Mean Accuracy: comparison of cross-encoder model trained on Devanagari vs SLP1}
    \label{tab:semantic-acc_dev_vs_slp1}
\end{table}

\begin{table}[ht]
    \centering
    
    \begin{tabular}{lcccc}
    \toprule
    \textbf{Model} & \multicolumn{2}{c}{\textbf{Raw}} & \multicolumn{2}{c}{\textbf{Isotonic Calibrated}} \\
    & Pearson & Spearman & Pearson & Spearman \\
    \midrule
    Cross-Encoder& 0.6767 & 0.5833 & 0.7248 & 0.7175 \\
    Cross-Encoder& 0.542 & 0.598 & 0.679 & 0.653 \\
    \bottomrule
    \end{tabular}
    
    \caption{Comparison of correlation with human semantic ratings (N=200) between cross-encoder trained using Sanskrit Devanagari vs SLP1 transliteration scheme}
    \label{tab:human_corrc_dev_vs_slp1}
\end{table}

\end{document}